\pdfoutput=1

\documentclass[11pt]{article}

\usepackage[]{ACL2023}

\usepackage{times}
\usepackage{latexsym}
\usepackage{amsmath}
\usepackage{adjustbox}
\usepackage{hyperref}

\usepackage[T1]{fontenc}

\usepackage[utf8]{inputenc}

\usepackage{microtype}

\usepackage{inconsolata}
\usepackage{makecell}

\usepackage{algorithm}
\usepackage{algorithmic}

\newcommand{\dataset}{\texttt{TempoWiC} }

\DeclareMathOperator*{\argmin}{arg\,min}
\DeclareMathOperator*{\argmax}{arg\,max}

\usepackage{listings}

\definecolor{codegreen}{rgb}{0,0.6,0}
\definecolor{codegray}{rgb}{0.5,0.5,0.5}
\definecolor{codepurple}{rgb}{0.58,0,0.82}
\definecolor{backcolour}{rgb}{0.95,0.95,0.92}

\lstdefinestyle{mystyle}{
    backgroundcolor=\color{backcolour},   
    commentstyle=\color{codegreen},
    keywordstyle=\color{magenta},
    numberstyle=\tiny\color{codegray},
    stringstyle=\color{codepurple},
    basicstyle=\ttfamily\footnotesize,
    breakatwhitespace=false,         
    breaklines=true,                 
    captionpos=b,                    
    keepspaces=true,                 
    numbers=left,                    
    numbersep=5pt,                  
    showspaces=false,                
    showstringspaces=false,
    showtabs=false,                  
    tabsize=4
}
\title{Large Language Models on Lexical Semantic Change Detection: \\ An Evaluation}

\author{Ruiyu Wang* \\
  University of Toronto \\
  \texttt{rywang@cs.toronto.edu} \\\And
  Matthew Choi* \\
  University of Toronto \\
  \texttt{mattchoi@cs.toronto.edu} \\}

\begin{document}
\maketitle
\def\thefootnote{*}\footnotetext{These authors contributed equally to this work.}\def\thefootnote{\arabic{footnote}}

\begin{abstract}
Lexical Semantic Change Detection stands out as one of the few areas where Large Language Models (LLMs) have not been extensively involved. Traditional methods like PPMI, and SGNS remain prevalent in research, alongside newer BERT-based approaches. Despite the comprehensive coverage of various natural language processing domains by LLMs, there is a notable scarcity of literature concerning their application in this specific realm. In this work, we seek to bridge this gap by introducing LLMs into the domain of Lexical Semantic Change Detection. Our work presents novel prompting solutions and a comprehensive evaluation that spans all three generations of language models, contributing to the exploration of LLMs in this research area.
\end{abstract}

\section{Introduction}
Languages are evolving. 
A perpetual flux of changes occurs, driven by an array of factors encompassing cultural, social, technological, and often, undiscovered influences. 
In this ever-shifting linguistic landscape, words shed unused senses while concurrently acquiring new meanings. Languages engage in a reciprocal exchange, borrowing senses from one another, and simultaneously exerting influence. This intricate web of linguistic evolution necessitates an automatic approach to comprehend and assess the fluidity of languages. Automation becomes the key to navigating and interpreting the dynamic currents of linguistic transformation.

However, the development of advanced computational methods for Diachronic Lexical Semantic Change (LSC) has been a blank slate for researchers. Since the 2010s, traditional embedding methods like PPMI, SVD, and SGNS have shown significant statistical correlation with human annotators and produced promising results in detecting shifts in word meaning \citep{kulkarni2014statistically, hamilton2018diachronic, schlechtweg_wind_2019}. As a result, previous works often lean towards using existing tools to uncover new meaning shifts rather than exploring novel algorithms to enhance them.
Additionally, frequency-based algorithms typically depend on large corpora \citep{tahmasebi2019survey}. Their performance on relatively low-resource datasets remains a challenge, and an efficient solution for this has yet to be discovered.

Since its introduction by \citet{vaswani2017attention}, models based on the Transformer architecture have become the latest trend. Contextualized word embeddings generated by BERT \citep{devlin2019bert} have provided a solid foundation for various downstream language tasks. Moreover, recently, Large Language Models (LLMs) have showcased remarkable capabilities in logical thinking and solving language tasks based on instructions \citep{bubeck2023sparks, zhao2023survey, openai_gpt-4_2023, roziere_code_2023}. This has inspired researchers to embrace LLMs for a modern approach to a series of lexical semantic tasks and explore their ability to understand natural language meanings.

In this study\footnote{Access the GitHub repository through this \href{https://github.com/ThisGuyIsNotAJumpingBear/AnotherWindOfChange}{\texttt{link}}.}, we conducted a series of tasks to assess the suitability of LLMs for LSC detection for \dataset \citep{loureiro_tempowic_2022}, a low-resource annotated tweet dataset. Our key findings are outlined as follows:
\begin{enumerate}
    \item We reassess the performance of \textbf{traditional methods} (i.e., PPMI, SGNS, SVD) in addressing diachronic LSC on a low-resource dataset.
    \item We introduce a simple yet innovative generative approach for diachronic LSC detection. This method achieves promising results without requiring fine-tuning on pre-trained models.
    \item We conduct comprehensive evaluations for LLMs, BERT-based methods, and traditional methods at both the corpus level and instance level, offering insights into their respective capabilities in diachronic LSC detection.
\end{enumerate}

\section{Related Works}
\label{sec:related-works}
\paragraph{Traditional Methods} We define traditional sense modeling methods as those that do not rely on pre-trained models and prior knowledge. These mainly include \textbf{count-based} vector representations and \textbf{predictive} vector representations. \textbf{Count- based} representations are computed based on word co-occurrence frequencies, with Positive Pointwise Mutual Information (PPMI) being a popular method. \textbf{Predictive} vector representations typically involve word2vec \citep{mikolov2013efficient} variants, such as the skip-gram with negative-sampling (SGNS) training method, which previously achieved state-of-the-art results on various tasks.

However, traditional methods have notable limitations. Firstly, due to their one-to-one relationship between words and vectors, these methods are inherently incapable of classifying colexification. Secondly, as all traditional models operate at the corpus level, they struggle to classify instance-level meaning changes, hindering in-depth analyses with human annotations produced for each instance. Thirdly, traditional models, particularly count-based ones, demand a large amount of training data to maintain performance, making the produced vectors susceptible to bias with a decreasing sample size.

\paragraph{Contextualized Representations} Contextualized representations are usually produced by BERT \citep{devlin2019bert} model or its variants. These methods produce word representations specific to the contexts of the word. Commonly, BERT representations under different contexts are clustered together and averaged to form the final representation of a word or a sense \citep{hu-etal-2019-diachronic, giulianelli-etal-2020-analysing}. Some practices also involve fine-tuning on corpora \citep{Martinc_2020}.

Contextualized representations effectively address issues raised by traditional methods. Concerning colexification, BERT embeddings can be clustered and averaged by senses, thereby producing sense representations instead of word representations \citep{hu-etal-2019-diachronic}. Additionally, as the original representations are not derived from the entire corpus but from individual input sequences, BERT supports in-depth meaning change analyses at the instance level and is robust to changes in the size of training data.

\paragraph{Generative Approaches}
almost all previous state-of-the-art models, including BERT, across various tasks. Models such as GPT-4 \citep{openai2023gpt4} and LLaMA \citep{touvron2023llama} have achieved near-human-level performance in question answering, reading comprehension, and logical reasoning. Recent works have shown that LLMs can excel in an even wider range of tasks with appropriate prompt engineering \citep{hou2023large, hegselmann2023tabllm}.

However, there are limited prior works on LLMs in the context of LSC detection. The most recent work on detecting lexical semantic changes uses a BART model \citep{lyu_mllabs-lig_2022}, which has significantly fewer trainable parameters compared to a modern LLM. Furthermore, we were unable to find any works that focus on a prompting paradigm and do not involve costly fine-tuning.

Although there are not many generative approaches in LSC detection for us to refer to, the remarkable capability of LLMs on unseen tasks still leads us to consider the possibility of employing a zero-shot generative approach for this task.

\section{Methodology}
We evaluate all three kinds of methods we mentioned in Section \ref{sec:related-works}. We present our selection of methods and descriptions for the novel approaches we made.

\subsection{Traditional Word Representations}
\paragraph{Positive Pointwise Mutual Information (PPMI)} is a word count matrix where each cell is weighted by the positive mutual information between the target word and the context. It is one of the most adapted count-based word representation methods. The transformed matrix is given by:
$$\scriptstyle M_{i,j}^{\texttt{PPMI}} = max\{log(\frac{\#(w_i, c_j) \sum_c (c)^\alpha}{\#(w_i) \#(c_j)^\alpha}) - log(k), 0\} $$
Where $w_i$ is the target word, $c_j$ is the context word, $k$ is the probability prior to observing $(w_i, c_j)$, and $\alpha$ is a smoothing factor to reduce bias caused by rare words. The representation matrix is counted in bigram, i.e. only adjacent words are counted as co-occurrence.

Since a count-based matrix is high-dimensional and sparse, we use truncated SVD to obtain a dimension-reduced approximation of PPMI. We have the dimensionality-reduced matrix $$M_{SVD} = U_k\Sigma_k$$where $U\Sigma V^T = M^{PPMI}$ and the top $k$ elements in $\Sigma$ is preserved, based on the eigenvalues.

\paragraph{Skip-gram with Negative Sampling (SGNS)}

SGNS follows the simple idea that words in the context should be close to one another. We can computationally follow this approach where for a vocabulary $V$, each word $w \in V$ and each context $c \in V$ can be represented by a $d$-dimensional vector by solving

$$\scriptstyle \argmax_\theta \sum_{(w,c) \in D} \log \sigma(v_c \cdot v_w) + \sum_{(w,c) \in D'} \log \sigma(-v_c \cdot v_w)$$

where $\sigma(x) = \dfrac{1}{1 + \epsilon^{-x}}$, $D$ is the set of all observed word-context-pairs and $D'$ is the set of randomly generated negative samples. $D'$ is created by drawing $k$ contexts from the unigram distribution $P(c) = \dfrac{\#(c)}{|D|}$ for each observation of $(w,c)$. it is generally noted that smaller datasets can afford to use a larger $k$ for more accurate results without incurring much computational cost. With the optimized parameters $\theta$ where $v_{c_i} = C_{i*}$ and $v_{w_i} = W_{i*}$ for $w,c \in V, i \in 1, ..., d$, we can get the final SGNS matrix $M$ by implicitly factorizing $M = WC^T$, and the $i$th row of $M$ will corresponds to the $w_i$'s $d$-dimensional semantic representation.

\subsubsection{Alignments}
\paragraph{Orthogonal Procrustes (OP)}
The orthogonal procrustes describes the least-square problem of approximating matrix $A$ to $B$ via an orthogonal transformation \citep{schonemann1966generalized}. By solving this problem on the two vector spaces yielded by the word representations, we can reach an optimum alignment between them after mapping.

We firstly create a binary matrix $D$ such that $D_{i, j} = 1$ if $w_i \in V_b$ and $w_J \in V_a$, where $V_a, V_b$ are the vocabularies for time periods $a, b$. One can also think of the matrix $D$ as the intersection of words between the time periods $a$ and $b$. The goal of OP is to find an optimal mapping matrix $W*$ such that the sum of the squared euclidean distance between $B_{i*}W$'s mapping and $A_{j*}$ for $D_{i, j}$ is minimized, in particular:

$$W^* = \argmin_W \sum_i \sum_j D_{i, j} ||B_{i*}W - A_{j*}||^2$$

We follow the procedure to mean-center, scale, and translate for the best results. The optimal solution is given by $W* = UV^T$ where $B^TDA = U \Sigma V^T$ is derived from the SVD of $B^TDA$. $A$ and $B$ are then aligned by $A^{OP} = A$ and $B^{OP} = BW*$. We use OP to align both PPMI and SGNS.

\subsubsection{Vector Distances}
\paragraph{Cosine Distance (CD)}
Cosine Distance is based on cosine similarity which measures the cosine of the angle of two non-zero vectors $\vec{x}$, $\vec{y}$ with equal magnitudes,
$$cos(\vec{x},\vec{y}) = \dfrac{\vec{x}\cdot\vec{y}}{\sqrt{\vec{x}\cdot \vec{x}}\sqrt{\vec{y}\cdot\vec{y}}}$$

cosine distance is then 

$$CD(\vec{x},\vec{y}) = 1 - cos(\vec{x},\vec{y})$$

\subsection{Contextualized Word Representations (BERT)}
As mentioned in Section \ref{sec:related-works}, BERT \citep{devlin2019bert} and its variants are the mostly adapted contextualized word representation models. 
In this work, we adopt a BERT (\texttt{bert-base-uncased}) model to generate contextualized word representations. It has 12 blocks with a hidden size of 768. The estimated parameter size is 110M. Given a size $n$ tokenized input sequence of $s_{in}=\{t_0, t_1, t_2, ..., t_n\}$, BERT produces an output sequence with the same length $s_{out} = \{v_0, v_1, ..., v_n\}$ where $v_i\in s_{out}$ is a vector of dimension 768 that encodes the representation of token on index $i$.

BERT uses WordPiece tokenization, which separates words into smaller tokens in some cases. The BERT representation does not change if the target word is not separated into word pieces. However, in the cases where it is separated, the representation of this word is defined as the average vector of all of its components. Formally, given an arbitrary word $w$ in sequence $s$ that can be separated into $m$ word pieces $\{p_1, p_2, ..., p_m\}$, the BERT representation vector $v_m$ is considered to be $$v_m = avg(\texttt{BERT}(p_1 | s), \texttt{BERT}(p_2 | s), ..., \texttt{BERT}(p_m | s))$$

The BERT vectors are used differently in each part of the evaluation. Detailed steps for utilizing the vectors for meaning shift detection are presented in relevant paragraphs under Section \ref{sec:corpus-level-lsc} and Section \ref{sec:instance-level-lsc}.

\subsection{LLM-based Methods}

We adopt the zero-shot prompting paradigm on an LLM. In detail, we present a pair of sentences that share a target word and ask if the meaning of the target word between the two tweets is different. By zero-shot, we converse with the LLM only once. LLM prompts usually allow two points of variation, the context for the LLM to have more information about the query, and the query itself. We carefully designed our prompts to include all necessary components of the context and the query. The detailed procedure can be found in algorithm \ref{algo:prompting}. The resulting prompt includes additional instructions on returning outputs. This ensures the response from the LLM returns only 0 or 1 and makes the computation of the evaluation metrics trivial. 


\begin{algorithm}
\caption{The prompting procedure.}
\label{algo:prompting}
\begin{algorithmic}
\FOR{data in dataset}
    \FOR{tweet1, tweet2, target\_word in data}
        \STATE p = prompt(tweet1, tweet2, target\_word)
        \STATE response = LLM(prompt)
        \STATE \# assertion: response $\in \{0, 1\}$
    \ENDFOR
\ENDFOR
\end{algorithmic}
\end{algorithm}

By using the above-mentioned steps to generate LLM responses on meaning shift, we are able to perform a set of analyses based on different task setups. Information on task-specific implementation is presented in Section \ref{sec:corpus-level-lsc} and Section \ref{sec:instance-level-lsc}. Some example prompts we used are presented in Appendix \ref{apdx:prompts}.

\section{Data and Task}
In this section, we present our dataset selection and choices of evaluation tasks. We adopt a tweet dataset with complete human annotations (Section \ref{sec:dataset}) and evaluate a model's performance on detecting both corpus-level meaning shifts (Section \ref{sec:corpus-level-lsc}) and instance-level meaning shifts (Section \ref{sec:instance-level-lsc}).
\subsection{Dataset Selection}
\label{sec:dataset}
We choose to use the \dataset dataset in our investigation. It consists of 3,287 sentence instances of 34 target words. The train/validation/test sets are separated in the original work. A full overview of dataset statistics can be found in Table \ref{tab:dataset-statistics}.

\begin{table}[]
    \centering
    \begin{tabular}{c|c|c}
         \textbf{Dataset} & \textbf{\#Target word} & \textbf{\#Instance} \\
         \hline
         Train & 15 & 1,428 \\
         Valid & 4 & 396 \\
         Test & 15 & 1,473 
    \end{tabular}
    \caption{Dataset statistics for \dataset. There are 34 target words with 3,287 instances under modeling.}
    \label{tab:dataset-statistics}
\end{table}

We use \dataset because of its unique characteristics. Firstly, it is a comparatively low-resource dataset compared to general lsc detection datasets. This gives us the ability to inspect the traditional methods' performance under this setup. Secondly, it provides both instance-level meaning shift annotation, and a corpus-level LSC evaluation, this provides us with sufficient resources for reference.

\subsection{Corpus-level LSC Detection} Detecting lexical semantic change on the corpus level is the most adapted computational approach. It requires computational methods to detect word changes in a holistic view. For the implementation of this task, we separate the \dataset dataset into clusters of years, generate appropriate measures for meaning change based on the representation methods, and determine the statistic correlation between the method predictions and the human annotations. 
\label{sec:corpus-level-lsc}
\paragraph{Traditional Models}
We generate the PPMI matrices for the three years 2019, 2020, and 2021. We experiment with different choices of $k$ for dimensionality reduction, ranging from $100$ to $500$. Similarly, We trained SGNS models for all 3 years. We then aligned each time frame (2019-2020, 2020-2021) with the orthogonal procrustes (mean-centered) for comparison. Lastly, we calculate the cosine distance between the overlapped target words in each time frame. We compile the cosine distance of all target words and calculate the pearson correlation to the annotated target word list, which is presented by table \ref{annotator-table}.

Detailed experiment results on different choices of hyperparameter are presented in Table \ref{sgns-table} and Table \ref{ppmi-table} in Appendix \ref{apdx:exp-result}.

\paragraph{BERT}
For each time frame and target word, we compute the average vector. Subsequently, we measure the lexical semantics change of each target word by comparing their average vectors in different years through cosine distance. We compile the cosine distance of all target words and calculate the pearson correlation to the annotated target word list.

\paragraph{LLM}
We adopt a \texttt{GPT-4}\footnote{Via OpenAI API: \texttt{GPT-4}.} model as our backbone. We crafted our prompts to always return binary values. To detect the corpus-level meaning change, we define a corpus-wise semantic change factor as an averaged pairwise identity score $$S = \frac{\# \texttt{LLM\_Choice}(S_i, S_j) = 1}{\# (S_i, S_j)}$$ where $S_i$ is an instance coming from time $i$, $S_j$ is an instance coming from time $j$, and \texttt{LLM\_Choice} is a function denoting the choice of LLM after receiving the prompted input. 
If the target word in $S_i$ and $S_j$ has the same meaning, then \texttt{LLM\_Choice}$=1$, denoting that there is no semantic change. 
We then calculate the pearson correlation with the human annotations.

\subsection{Instance-level Meaning Shift Detection} Since \dataset provides pairs of tweets with human-annotated ground truths, we also attempt to examine different models' capabilities of distinguishing meanings in those pairs. Given the two tweets posted in different periods and the target word, the model is required to predict either 0 (target word meaning changed) or 1 (target word meaning unchanged). We yield an accuracy score and an f1 score after comparing the model performance with the ground truths.
\label{sec:instance-level-lsc}
\paragraph{Traditional Methods}
As discussed in Section \ref{sec:related-works}, one of the major restrictions of the traditional methods is that they are not able to catch individual instance-level semantic shifts. As a result, we have to exclude this set of models from our evaluation of this task.

\paragraph{BERT}
In contrast to the traditional word representation methods, BERT produces an individual embedding for each input word, which enables BERT-based methods to perform instance-level meaning change detection. For each tweet pair $t_1, t_2$ provided by \dataset, we generate two representations for target word $w$, i.e. $$v_1 = \texttt{BERT}(w | t_1), v_2 = \texttt{BERT}(w | t_2)$$ and compute their cosine distance as $d = CD(v_1, v_2)$. We then optimize a threshold for the best meaning change classification and calculate the accuracy and f1 score based on the provided ground truth. 

\paragraph{LLM}
Similarly to corpus-level detection, we adopt a \texttt{GPT-4} model as our backbone and determine the semantic change within two sentences for a target word. Based on the LLM response, we calculate the f1 score and accuracy by comparing it with the ground truths provided.

\section{Results}

\begin{figure*}[h!]
    \centering
    \includegraphics[width=0.8\linewidth]{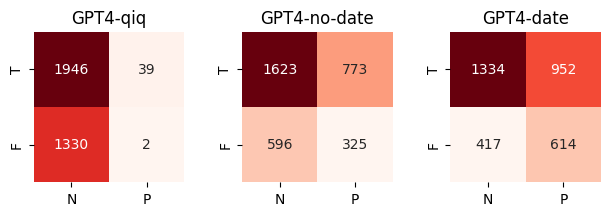}
    \caption{Confusion Matrices for different prompts with GPT4. Suffix `date' includes the data information in the prompt, and `no-date' does not include such information. `qiq' (question-in-query) is the prompt that includes the date, but uses a question for LLM response instead of an instruction. T/F stands for True/False. N/P stands for negative/positive. While \texttt{GPT4-date} has the best prediction result presented by the its confusion matrix, \texttt{GPT4-qiq} appears to have a very low number of true positives, which suggests the underlying cause of its low F1 score, presented in Table \ref{instance-metric-table}.}
    \label{con_mat}
\end{figure*} 

\subsection{Corpus-level LSC Detection Methods}

\begin{table}[h!]
\centering
\begin{adjustbox}{max width=0.48\textwidth}
 \begin{tabular}{|c|c|c|} 
 \hline
 \textbf{Model} & \textbf{r} & \textbf{p-value} \\
 \hline
 PPMI(dim=300) & -0.005 & 0.427 \\
 SGNS(k=15, vec\_size=100) & 0.47 & 0.05 \\
 BERT & 0.63 & 0.00675 \\
 \textbf{GPT-4(no-date)} & \textbf{-0.66} & \textbf{2.12e-5} \\
 \hline
 \end{tabular}
 \end{adjustbox}
 \caption{\label{corpus-pearson-table} The pearson correlations and their respective p-values for all models. The result is chosen from each model with the best set of hyperparameters. If there are multiple time frames (i.e. separated results for 2019-2020 and 2020-2021) in calculating the p-values, the average is taken. GPT-4 (with no date information included during prompting) had the best pearson correlation and the p-value.}
\end{table}

We report the pearson correlation and their respective p-value for the best-performing models from our hyperparameters in Table \ref{corpus-pearson-table}. GPT-4 with no date information appears to be the best model among all the candidates, followed by BERT. The traditional models perform the worst, suggesting that they are not capable of evaluating lexical semantic changes in this resource-low dataset. GPT-4 outperforms BERT with a 4\% absolute improvement in pearson correlation to the annotated labels.

We note a strong negative correlation for GPT-4. The annotator labels provide percentages for the amount of change for a particular target word. For example, 0.52 would mean that 52\% of all tweet pairs signified the meaning of the target word has changed and a larger coefficient would mean more meaning change. However, the labels in \dataset are presented differently. They are used to represent a True/False(1/0) predicate for the question "Is the meaning of the target word in the 2 tweets the same?". For example, a label of False(0) would signify the meaning of the target word is indeed different. When calculating the average meaning change for a target word, a smaller value relates to more meaning change. Hence, we verify that the negative correlation is an expected outcome.  

\subsection{Instance-level Meaning Change Methods}
We present a more detailed inspection of the BERT and LLM-based methods since the traditional methods fail to perform this task. The result is displayed in Table \ref{instance-metric-table}.
\begin{table}[h!]
\centering
\begin{adjustbox}{max width=0.48\textwidth}
 \begin{tabular}{|c|c|c|} 
 \hline
 \textbf{Model} & \textbf{F1 Score} & \textbf{Accuracy} \\
 \hline
 BERT (t=0.4) & 0.62 & 0.46 \\
 BERT (t=0.6) & 0.62 & 0.53 \\
 BERT (t=0.7) & 0.60 & 0.64 \\
 GPT-4 (qiq) & 0.06 & 0.60 \\
 GPT-4 (date) & \textbf{0.65} & 0.69 \\
 GPT (no-date) & 0.63 & \textbf{0.72} \\
 \hline
 \end{tabular}
 \end{adjustbox}
 \caption{\label{instance-metric-table} F1 scores and Accuracy for the top 3 BERT models and GPT-4 with all prompting styles. The values of $t$ in the BERT models refer to the threshold choices. For GPT-4 prompts, `date' includes the data information in the prompt, and `no-date' does not include such information. `qiq' is the prompt that includes the date, but uses a question for LLM response instead of an instruction. Prompting details can be found in the Appendix.}
\end{table}
The accuracy scores suggest an observable gap between GPT-4 and BERT. GPT-4 is the only model that reaches the accuracy of 70\%, and using different prompts does not significantly reduce the performance, which indicates the robustness of setups. GPT-4 surpasses the best performing BERT model by 8\% in F1 score and 11\% in accuracy.

Viewing the F1 scores, we found that the performance of the models is close to each other except for GPT-4 with question-asking prompts. This refers to the importance of prompt engineering. When the question "Is the meaning of \{target word\} different in the two tweets" was asked in the context instead of an instruction, it produced night and day results. While accuracy is better, the F1 score between these two prompts reports a significant difference. We investigated further by plotting the confusion matrix for a clearer picture of this discrepancy. As Figure \ref{con_mat} illustrates, the question-in-query (qiq) format had a very low number of true positives, which caused the f1 score to be extremely low. Moreover, We found out that \dataset had at least 60\% of data without meaning changes, leading to a decent accuracy even if GPT-4 is prompted to predict "no change" every time. Hence, We decided to use the F1 score as the main predictor of the model's performance, as it produces insights to the model performance that accuracy does not. Following that decision, the best GPT-4, with a carefully designed prompt, still performs better across the board compared to BERT.

\section{Discussion}
During the performance of our evaluation tasks, we made several key observations as follows:

\paragraph{The outstanding performance of LLMs} Apparently, GPT-4 is consistently outperforming other methods. While it is a rather trivial question to ask why it performs better than the other models, we hypothesize that GPT-4 being a 175 billion parameter model and trained on Petabytes of data does make it excellent at different language tasks, and this bleeds over into the realm of semantic change. 

\paragraph{Causes of traditional model's failure} Another intriguing question would be why the traditional methods such as PPMI and SGNS perform so poorly, not only compared to the new models we bring into comparison on \dataset, bu also to its good performance for more traditional LSC datasets such as n-gram google books as \citet{schlechtweg_wind_2019} reports. As discussed in Section \ref{sec:related-works}, we believe that the cause lies in the size of the dataset. We hypothesize that the traditional methods require a much larger corpus, in general, to take significant variations into account to count for PPMI and contexts for SGNS to perform well.

Moreover, we assert that the narrow time frame is not a contributing factor to the subpar performance of traditional methods. The selected target words were "trending" during that specific time frame. This implies that if a tweet pair exhibits a change in meaning for a target word, it is highly probable that the change is substantial, providing sufficient room for fault tolerance.

\paragraph{A global comparison} Another advantage of choosing to utilize \dataset is that it is an open competition with a leaderboard, allowing for a global comparison with other baseline models. Based on the results from the \dataset competition provided by \citet{lyu_mllabs-lig_2022}, our zero-shot prompting outperforms a BART-based method that employs a similar format to GPT4-qiq but fine-tunes the backbone instead. Achieving the current F1 score and accuracy without the need for additional resources for fine-tuning not only represents significant progress in this task but also showcases the capability of LLMs to comprehend natural language inputs. Given this commendable performance, we believe it would be worthwhile to investigate the effectiveness of in-depth GPT prompt engineering on computational methods for semantic change as a low-effort benchmark for future research.

\section{Conclusion}
In this study, we introduce a zero-shot solution based on large language models for detecting both corpus-level and instance-level semantic changes. Through a comprehensive evaluation on the human-annotated tweet dataset \dataset, we demonstrate the advantages of our approach in all setups, surpassing BERT in both tasks. Additionally, we reveal existing issues with traditional methods (e.g., PPMI, SGNS) and find that our method, relying solely on zero-shot LLM inference, stands among the top-performing models globally. We anticipate that, with further advancements and exploration in generative modeling, LLMs could become an efficient tool in this research domain.

\bibliography{anthology,custom}
\bibliographystyle{acl_natbib}

\appendix

\section{LLM Prompts}

\begin{table}[h!]
\centering
 \begin{tabular}{|c c|} 
 \hline
 Word & Meaning change \\
 \hline
 frisk & 0.54 \\
 pogrom & 0.05 \\
 containment & 0.33 \\
 virus & 0.48 \\
 epicenter & 0.71 \\
 ventilator & 0.17 \\
 villager & 0.64 \\
 turnip & 0.95 \\
 bunker & 0.61 \\ 
 mask & 0.76 \\
 teargas & 0.03 \\
 paternity & 0.22 \\
 entanglement & 0.89 \\
 folklore & 0.92 \\
 parasol & 0.85 \\
 impostor & 0.76 \\
 lotte & 0.43 \\
 recount & 0.28 \\
 primo & 0.77 \\
 milker & 0.50 \\
 moxie & 0.83 \\
 unlabeled & 0.90 \\
 pyre & 0.32 \\
 gaza & 0.60 \\
 ido & 0.83 \\
 airdrop & 0.40 \\
 bullpen & 0.09 \\
 crt & 0.68 \\
 monet & 0.94 \\
 burnham & 0.16 \\
 delta & 1.00 \\
 gala & 0.46 \\
launchpad & 0.81 \\
vanguard & 0.95 \\
 \hline
 \end{tabular}
 \caption{\label{annotator-table} All words available in the test set and their respective annotator labelled percent meaning change. The value is scaled to 0 to 1 with 1 meaning a larger amount of meaning change for that particular target word.}
\end{table}

\label{apdx:prompts}
Listing 1, 2, and 3 present the examples of different prompts used in different setups. 

\begin{lstlisting}[caption={GPT4-no-date context and query.}, language=python]
gpt4_no_date_context = ''' 
    Given two tweets, their dates and a target word in the form of "### Tweet1: ... Date: ... ### Tweet2: ... Date: ... ### Target Word: ...", 
    Tell me whether the meaning of the target word is different in the two tweets. Furthermore, only respond with 0 if it is a yes and 1 if it is a no. Do not explain.
'''

gpt4_no_date_query = '''
### Tweet1: {tweet1} ### Tweet2: {tweet2} ### Target Word: {word}
'''
\end{lstlisting}

\begin{lstlisting}[caption={GPT4-date context and query}, language=python]
gpt4_date_context = ''' 
    Given two tweets, their dates and a target word in the form of "### Tweet1: ... Date: ... ### Tweet2: ... Date: ... ### Target Word: ...", 
    Tell me whether the meaning of the target word is different in the two tweets. Furthermore, only respond with 0 if it is a yes and 1 if it is a no. Do not explain.
'''

gpt4_date_query = '''
### Tweet1: {tweet1} Date: {date1} ### Tweet2: {tweet2} Date: {date2} ### Target Word: {word}
'''
\end{lstlisting}

\begin{lstlisting}[caption={GPT4-qiq context and query}, language=python]
gpt4_qiq_context ='''
You are given two tweets with their respective dates of creation and a question in the format of Tweet-1: ... Tweet-2: ... Question: ...
Answer the question with 0 if it is a yes and 1 if it is a no. Do not explain.
'''

gpt4_qiq_query = '''
Tweet-1: {tweet1} Date: {date1} Tweet-2: {tweet2} Date: {date2} Question: Is the meaning of {word} different in the last 2 tweets?
'''
\end{lstlisting}

\section{Experiment Results of Different Hyperparameter Choices}
\label{apdx:exp-result}
Table \ref{sgns-table} presents the result of top 5 SGNS models, their corresponding hyperparameter choices and pearson correlation on different frames. Table \ref{ppmi-table} presents the result of all ppmi models with different choice of reduced dimensionality. The average pearson correlation and p-values across frames of years are presented.

\begin{table}[h!]
\centering
\begin{adjustbox}{max width=0.48\textwidth}
 \begin{tabular}{|c c c c c|} 
 \hline
 k & vector size & r(2019-2020) & r(2020-2021) & \textbf{avg r} \\
 \hline
 10 & 100 & 0.53* & 0.21 & \textbf{0.37} \\
 15 & 100 & 0.39 & 0.31 & \textbf{0.36} \\
 12 & 100 & 0.46* & 0.25 & \textbf{0.35} \\
 5 & 100 & 0.30 & 0.27 & \textbf{0.28} \\
 10 & 200 & 0.36 & 0.10 & \textbf{0.23} \\
 \hline
 \end{tabular}
 \end{adjustbox}
 \caption{\label{sgns-table} The top 5 SGNS models in terms of pearson correlation to the annotator labels. * represents $p<0.05$}
\end{table}

\begin{table}[h!]
\centering
\begin{adjustbox}{max width=0.7\textwidth}
 \begin{tabular}{|c c c|} 
 \hline
 Dimension  & avg r & avg p-value \\
 \hline
100 & -0.034 & 0.626 \\
200 & -0.024 & 0.586 \\
300 & \textbf{-0.005} & \textbf{0.427} \\
400 & -0.002 & 0.570 \\
500 & -0.029 & 0.670 \\
 \hline
 \end{tabular}
 \end{adjustbox}
 \caption{\label{ppmi-table} The results of PPMI models reduced to different dimensionalities in terms of pearson correlation and p-value to the annotator labels. Results with the lowest p-value is highlighted.}
\end{table}

\end{document}